\documentclass{article}
\usepackage{spconf,amsmath,graphicx}
\usepackage{algorithm,algorithmic,cite,amsmath,amssymb,epsfig,enumerate,amsfonts,multirow,array,tabularx,booktabs,stfloats,flushend}
\usepackage[T1]{fontenc}

\usepackage[colorlinks=true, linkcolor=black, anchorcolor=black, citecolor=black]{hyperref}

\usepackage{array}
\newcolumntype{I}{!{\vrule width 3pt}}
\newlength\savedwidth

\newlength\savewidth

\def\x{{\mathbf x}}

\def\n{{\mathbf n}}
\def\f{{\mathbf f}}

\def\x{{\mathbf x}}

\title{Multilayer bootstrap network for unsupervised speaker recognition}
\name{Xiao-Lei~Zhang}
\address{Department of Computer Science and Engineering, The Ohio State University, Columbus, OH, USA\\xiaolei.zhang9@gmail.com}
\begin{document}
%
\maketitle
\begin{abstract}
We apply multilayer bootstrap network (MBN), a recent proposed unsupervised learning method, to unsupervised speaker recognition. The proposed method first extracts supervectors from an unsupervised universal background model, then reduces the dimension of the high-dimensional supervectors by multilayer bootstrap network, and finally conducts unsupervised speaker recognition by clustering the low-dimensional data. The comparison results with 2 unsupervised and 1 supervised speaker recognition techniques demonstrate the effectiveness and robustness of the proposed method.
\end{abstract}
\begin{keywords}
multilayer bootstrap network, speaker recognition, unsupervised learning.
\end{keywords}
\section{Introduction}
\label{sec:intro}

Speaker recognition aims to identify speakers from their voices. It is important in many speech systems, such as speaker diarization, language recognition, and speech recognition. Supervised methods include maximum a posteriori estimation \cite{reynolds2000speaker,campbell2006svm}, linear discriminative analysis (LDA) \cite{kenny2007joint,dehak2011front}, support vector machines \cite{campbell2006svm}, deep neural networks \cite{chen2011learning,zhao2015cochannel}, etc.


Because constructing a manually-labeled corpus is laboring intensive and time-consuming, it is strongly needed to develop unsupervised speaker recognition methods. Existing methods mainly include principle component analysis (PCA), $k$-means clustering, Gaussian mixture model (GMM), agglomerative hierarchical clustering, and joint factor analysis. For example, Wooters and Huijbregts \cite{wooters2008icsi} used agglomerative clustering to merge speaker segments by Bayesian information criterion. Iso  \cite{iso2010speaker} used vector quantization to encode speech segments and used spectral clustering, which is a $k$-means clustering applied to a low-dimensional subspace of data, for speaker recognition.
Nwe \textit{et al.} \cite{nwe2012speaker} used a group of GMM clusterings to improve individual base GMM clusterings. Some methods apply clustering techniques, e.g. variational Bayesian expectation-maximization (EM) GMM \cite{shum2013unsupervised} and spectral clustering \cite{tawara2015comparative}, to a low-dimensional \textit{total variability subspace} \cite{dehak2011front} that is learned from high-dimensional supervectors by joint factor analysis \cite{dehak2011front}. Some methods compensate the total variability space with new items, e.g. \cite{wu2012intra}.

Because little prior knowledge of data is known beforehand, an unsupervised method should satisfy the following conditions: (i) no need for manually-labeled training data; (ii) no hyperparameter tunning for a satisfied performance;  and (iii) robustness to different data or modeling conditions. Due to these strict requirements, unsupervised speaker recognition is a very difficult task. In this paper, we present a multilayer bootstrap network (MBN) \cite{zhang2014nonlinear} based algorithm. MBN is a recent proposed unsupervised nonlinear dimensionality reduction algorithm. Experimental results show that the proposed method satisfies these requirements.

This paper is organized as follows. In Section \ref{sec:system}, we present the MBN-based system. In Section \ref{sec:mbn}, we present the MBN algorithm and its typical hyperparameter setting. In Section \ref{sec:related_work}, we present the relationship between MBN and deep learning. In Section \ref{sec:experiment}, we report comparison results. In Section \ref{sec:conclusion}, we conclude this paper.

\section{System}\label{sec:system}
Given an unlabeled speaker recognition corpus, we propose the following unsupervised algorithm:\footnote{The source code is downloadable from {http://sites.google.com/site/ zhangxiaolei321/speaker\_recognition}}
\begin{itemize}
\itemsep=0.0pt
  \item The first step trains a speaker- and session-independent unsupervised universal background model (UBM) \cite{reynolds2000speaker} from an acoustic feature, which produces a $d$-dimensional supervector for each utterance, denoted as $\x=[\n^T,\f^T]^T$ where $\n$ is the accumulation of the mixture occupation over all frames of the utterance and $\f$ is the vector form of the centered first order statistics.

    \item The second step reduces the dimension of $\x$ from $d$ to $\bar{d}$ ($\bar{d}\ll d$) by multilayer bootstrap network (MBN) which is introduced in Section \ref{sec:format}.

    \item The third step conducts $k$-means clustering on the low-dimensional data if the number of the underlying speakers is known, or agglomerative clustering if the number of the speakers is unknown.
\end{itemize}

\section{Multilayer bootstrap network}\label{sec:mbn}
\label{sec:format}

The structure of MBN \cite{zhang2014nonlinear} is shown in Fig. \ref{fig:0}.
 MBN is a multilayer localized PCA algorithm that gradually enlarges the area of a local region implicitly from the bottom hidden layer to the top hidden layer by high-dimensional sparse coding, and gets a low-dimensional feature explicitly by PCA at the output layer.

Each hidden layer of MBN consists of a group of mutually independent $k$-centers clusterings. Each $k$-centers clustering has $k$ output units, each of which indicates one cluster. The output units of all clusterings are concatenated as the input of their upper layer \cite{zhang2014nonlinear}.

 MBN is trained layer-by-layer from bottom up. For training a hidden layer given a $d$-dimensional input $\mathcal{X} = \left\{\mathbf{x}_1,\ldots,\mathbf{x}_n \right\}$, MBN trains each clustering independently  \cite{zhang2014nonlinear}:

  \begin{figure}[t]
 \centering
         \includegraphics[width=8.25cm]{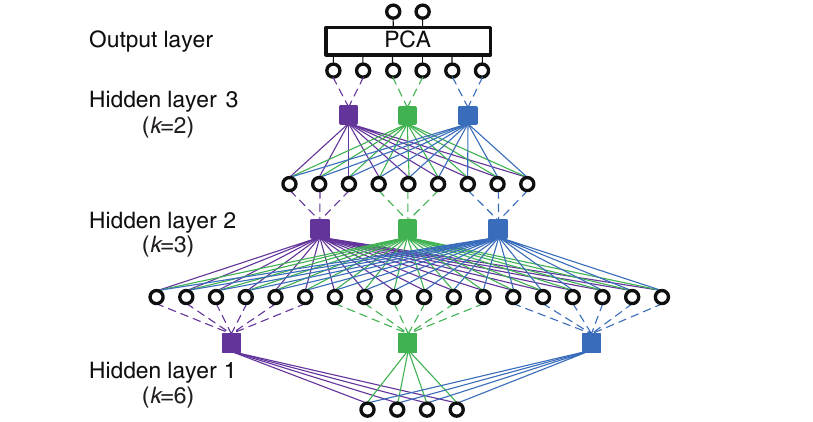}
 \caption{{The MBN network.} Each square represents a $k$-centers clustering.}
  \label{fig:0}
 \end{figure}

\begin{figure}[t]
 \centering
         \includegraphics[width=5.195cm]{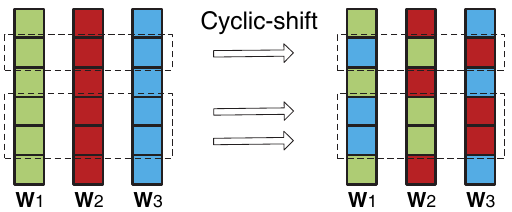}
 \caption{{Random reconstruction step in MBN.}}
  \label{fig:0_2}
 \end{figure}


 \begin{itemize}
 \itemsep=0.0pt
   \item \textbf{Random feature selection.} The first step randomly selects $\hat{d}$ dimensions of $\mathcal{X}$ ($\hat{d}\le d$) to form a new set $\hat{\mathcal{X}} = \left\{\hat{\mathbf{x}}_1,\ldots,\hat{\mathbf{x}}_n \right\}$. This step is controlled by a hyperparameter $a=\hat{d}/d$.
   \item \textbf{Random sampling.} The second step randomly selects $k$ data points from $\hat{\mathcal{X}}$ as the $k$ centers of the clustering, denoted as $\{\mathbf{w}_{1},\ldots,\mathbf{w}_{k} \}$. This step is controlled by a hyperparameter $k$.
    \item \textbf{Random reconstruction.} The third step randomly selects $d'$ dimensions of the $k$ centers ($d'\le \hat{d}/2$) and does a one-step cyclic-shift as shown in Fig. \ref{fig:0_2}. This step is controlled by a hyperparameter $r=d'/\hat{d}$.
   \item \textbf{Sparse representation learning.} The fourth step assigns the input $\hat{\mathbf{x}}$ to one of the $k$ clusters and outputs a $k$-dimensional indicator vector $\mathbf{h} = [h_1,\ldots,h_k]^T$. For example, if $\hat{\mathbf{x}}$ is assigned to the second cluster, then $\mathbf{h} = [0,1,0,\ldots,0]^T$. The assignment is calculated according to the similarities between $\hat{\mathbf{x}}$ and the $k$ centers,
 in terms of some
 predefined similarity measurement at the bottom layer, such as the minimum squared loss $ \arg\min_{i=1}^{k}\|\mathbf{w}_i-\hat{\mathbf{x}}\|^2$, or in terms of $\arg\max_{i=1}^{k}\mathbf{w}_i^T\hat{\mathbf{x}}$ at all other hidden layers \cite{zhang2014nonlinear}.
 \end{itemize}

 \subsection{A typical hyperparameter setting}
 MBN has five hyperparameters $\left\{V,L,\{k_l\}_{l=1}^L,a,r \right\}$ where $V$ is the number of $k$-centers clusterings per layer, $L$ is the number of hidden layers, and $k_l$ is the hyperparameter $k$ at the $l$th hidden layer. As shown in \cite{zhang2014nonlinear}, MBN is robust to hyperparameter selection. Here we introduce a typical setting:

 \begin{itemize}
 \itemsep=0.0pt
   \item \textbf{Setting hyperparameter $k$.} (i) $k_1$ should be as large as possible, i.e. $k_1\rightarrow n$. Suppose the largest $k$ supported by hardware is $k_{\max}$, then $k_1 = \min(0.9n, k_{\max})$. (ii) $k_l$ decays with a factor of, e.g. 0.5, with the increase of hidden layers. That is to say, $k_l=0.5k_{l-1}$. (iii) $k_L$ should be larger than the number of speakers $c$. Typically, $k_L \approx1.5c$. If $c$ is unknown, we simply set $k_L$ to a relatively large number, e.g. 30, since $c$ is unlikely larger than 30 in a practical dialog.
   \item \textbf{Setting hyperparameter $r$.} When a problem is small-scale, e.g. $k_1 > 0.8n$, then $r=0.5$; otherwise, $r=0$.
    \item \textbf{Setting other hyperparameters.} Hyperparameter $V$ should be at least larger than 100, typically $V = 400$. Hyperparameter $a$ is fixed to 0.5. Hyperparameter $L$ is determined by $k$.
 \end{itemize}

\section{Related work}\label{sec:related_work}

The proposed method learns multilayer nonlinear transforms, which is related to deep learning (a.k.a., multilayer neural networks)---a recent advanced topic in many speech processing fields, e.g.  speaker recognition \cite{chen2011learning,zhao2015cochannel}, speech recognition \cite{dahl2012context}, speech separation and enhancement \cite{wang2013towards,xu2015regression,huang2015joint,zhang2015deep}, speech synthesis \cite{ling2013modeling}, and voice activity detection \cite{zhang2013deep,zhang2015boosting}. The aforementioned deep learning methods are all supervised ones and limited to neural networks, while the proposed method is an unsupervised one and different from neural networks.



\section{Experiments}\label{sec:experiment}

\subsection{Experimental setup}
We used the training corpus of speech separation challenge (SSC) \cite{SSC}. The training corpus contains 34 speakers, each of which has 500 clean utterances. We selected the first 100 utterances (a.k.a, sessions) of each speaker for evaluation, which amounts to 3400 utterances. We set the frame length to 25 milliseconds and frame shift to 10 milliseconds, and extract a $25$-dimensional MFCC feature.

For the proposed MBN-based speaker recognition, we adopted the typical parameter setting of MBN. Specifically, $V=400$, $a=0.5$, $r=0.5$, and $k$ were set to $3060$-$1530$-$765$-$382$-$191$-$95$. The output of PCA was set to $\{2,3,5,10,30,50\}$ dimensions respectively. We assumed that the number of speakers was known, and used $k$-means clustering for clustering the low-dimensional data.

We compared with PCA, $k$-means clustering, and an LDA-based system, where the first two methods are unsupervised and the third one is supervised. For the PCA-based method, we first used the same UBM as the MBN-based method to extract high-dimensional supervectors, then reduced the dimension of the supervectors to $\{2,3,5,10,30,50\}$ respectively, and finally evaluated the low-dimensional output of PCA by $k$-means clustering. For the $k$-means-clustering-based method, we apply $k$-means clustering to the high-dimensional supervectors directly.

The LDA-based system\footnote{The source code is downloadable from http://research.microsoft.com/en-us/downloads/a6262fec-03a7-4060-a08c-0b0d037a3f5b/} uses UBM to extract a high-dimensional feature, then uses {joint factor analysis} to reduce the high-dimensional feature to an intermediately low dimensional representation in an unsupervised way, and finally uses LDA, a supervised dimensionality reduction method, to reduce the intermediate representation to a low-dimensional subspace where classification is conducted by a probabilistic LDA algorithm. Since factor analysis is an unsupervised dimensionality reduction method, we set its output to $\{2,3,5,10,30,50\}$ dimensions respectively for comparison. We constructed a training set from the SSC corpus for this supervised method: each speaker consists of 100 training utterances, which are selected from the 400 remaining utterances of the speaker.

The performance was measured by normalized mutual information (NMI) \cite{strehl2003cluster}. MNI was proposed to overcome the label indexing problem between the ground-truth labels and the predicted labels. It is one of the standard evaluation metrics of unsupervised learning. The higher the NMI is, the better the performance is. We also report the classification accuracy of the LDA-based system in the {Supplementary Material}\footnote{http://sites.google.com/site/ zhangxiaolei321/speaker\_recognition} where we can see that NMI is consistent with classification accuracy.

\subsection{Results}

Because all comparison methods use UBM to extract speaker- and session-independent supervectors, we need to study how they behave in different UBM settings, in terms of mixture number and expectation-maximization (EM) iterations. (i) The mixture number of UBM reflects the capacity of UBM for modelling an underlying data distribution: if the mixture number of UBM is smaller than the number of speakers, UBM is likely \textit{underfitting}, i.e. it cannot grasp the data distribution well. To study this effect, we set the mixture number of UBM to $\{1,2,4,8,16,32,64\}$ respectively. (ii) The number of EM iterations of UBM reflects the quality of the acoustic feature produced by UBM: if the EM optimization is not sufficient, the acoustic feature is noisy. To study this effect, we set the number of EM iterations of UBM to $\{0, 20\}$ respectively, where setting the number of iterations to $0$ means that UBM is initialized with randomly sampled means without EM optimization, which is the worst case.
\begin{figure}[t]
 \centering
         \includegraphics[width=8.25cm]{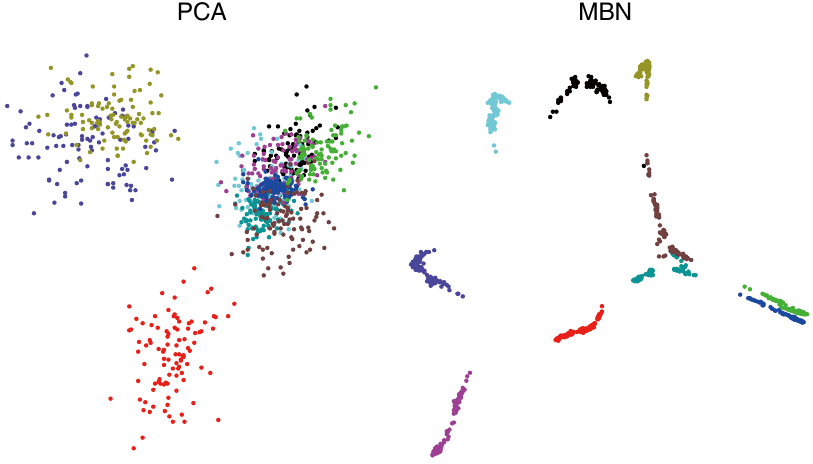}
 \caption{{Visualizations of 10 speakers by PCA and MBN respectively, where a 16-mixtures UBM with 20 EM iterations is used to produce their input supervectors. The speakers are labeled in different colors.}}
  \label{fig:1}
 \end{figure}

     \begin{figure}[t]
 \centering
         \includegraphics[width=8.25cm]{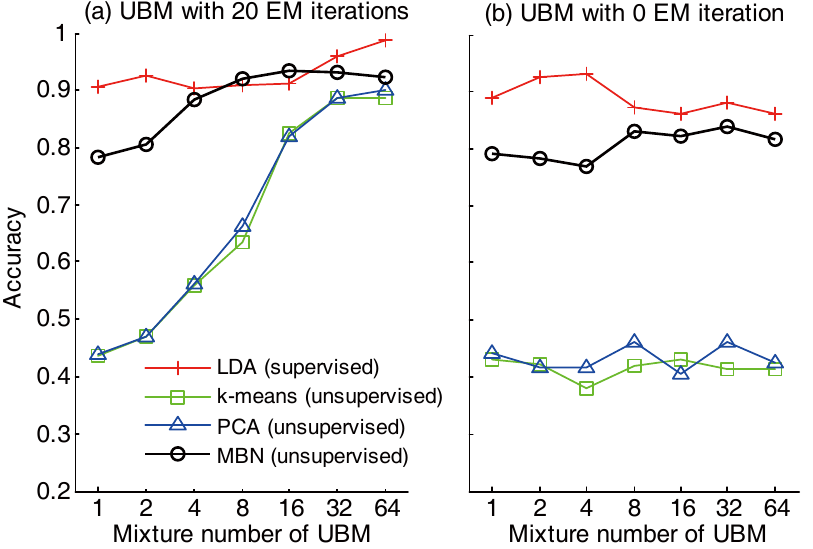}
 \caption{{Accuracy comparison (in terms of NMI) between LDA-, $k$-means clustering-, PCA-, and MBN-based speaker recognition methods with respect to the mixture number of UBM. (a) Comparison when the EM iteration number of UBM is set to $20$. (b) Comparison when the EM iteration number of UBM is set to $0$. Note that given a mixture number of UBM, the accuracy of a method is the best result among the results produced from 6 candidate output dimensions of the method, except $k$-means clustering.}}
  \label{fig:2}
 \end{figure}

   \begin{figure}[t]
 \centering
         \includegraphics[width=8.25cm]{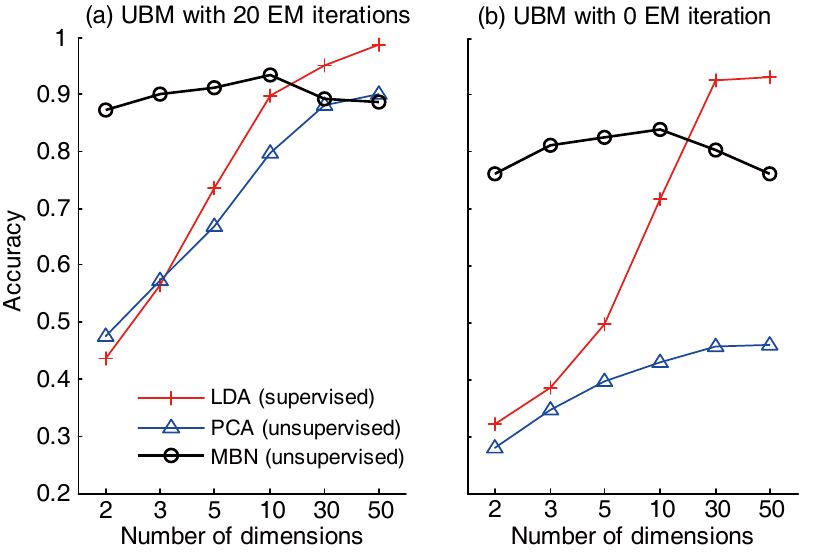}
 \caption{{Accuracy comparison (in terms of NMI) between LDA-, PCA-, and MBN-based speaker recognition methods with respect to the number of output dimensions. (a) Comparison when the EM iteration number of UBM is set to $20$. (b) Comparison when the EM iteration number of UBM is set to $0$. Note that given a number of output dimensions, the accuracy of a method is the best result among the results produced from 7 candidate UBMs.}}
  \label{fig:3}
 \end{figure}

    \begin{figure*}[t]
 \centering
         \includegraphics[width=17.8cm]{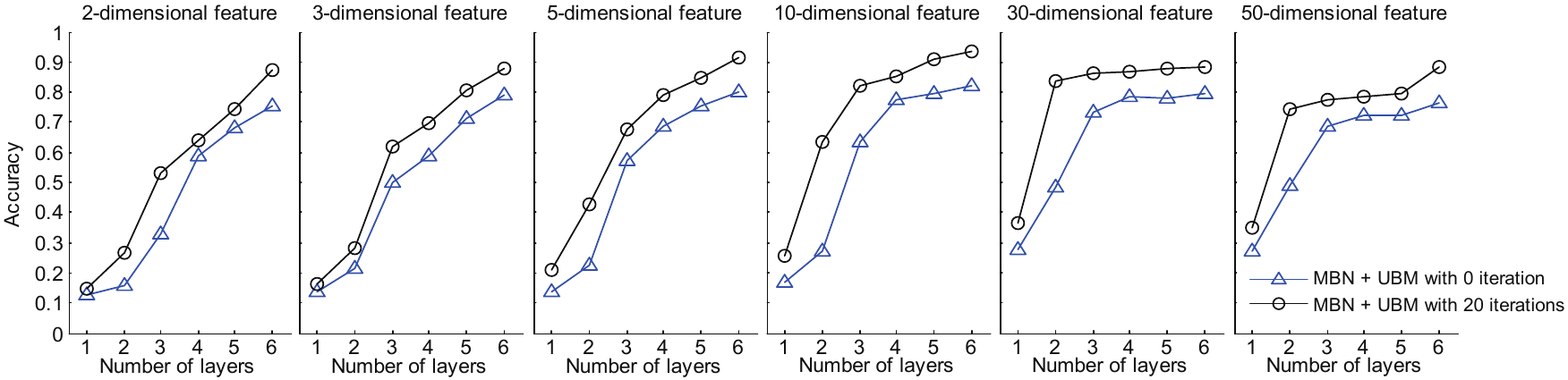}
 \caption{{Accuracy (in terms of NMI) of MBN-based method with respect to the number of hidden layers.}}
  \label{fig:4}
 \end{figure*}

Fig. \ref{fig:1} and Supplementary-Fig. 1 give a comparison example between PCA and MBN in visualizing the first 10 speakers, where a 16-mixtures UBM with 20 and 0 EM iteration are used to generate their inputs respectively. From the figures, we can see that MBN produces ideal visualizations.

Fig. \ref{fig:2} reports results with respect to the mixture number of UBM. Fig. \ref{fig:3} reports results with respect to the number of output dimensions. Supplementary-Tables 1 and 3 report the detailed results of the two figures. From the figures and tables, we observe the following phenomena: (i) the MBN-based method outperforms the PCA- and $k$-means-clustering-based methods and approaches to the supervised LDA system in all cases; (ii) the MBN-based method is less sensitive to different parameter settings of both UBM and MBN itself; (iii) the LDA-based system is less sensitive to the mixture number of UBM, but sensitive to the number of output dimensions; (iv) the PCA-based method is sensitive to both the mixture number of UBM and the number of output dimensions, and strongly relies on the effectiveness of UBM; (v) the performance of the $k$-means-clustering-based method is consistent with that of the PCA-based method.

Fig. \ref{fig:4} reports results of the MBN-based method with respect to the number of hidden layers. From the figure, we observe that the accuracy improves gradually with the increase of the number of hidden layers.








\section{conclusions}\label{sec:conclusion}
\label{sec:majhead}

In this paper, we have proposed a multilayer bootstrap network based unsupervised speaker recognition algorithm. The method first uses UBM to extract a high-dimensional feature from the original MFCC acoustic feature, then uses MBN to reduce the high-dimensional feature to a low-dimensional space, and finally clustering the low-dimensional data. We have compared it with the PCA-, $k$-means-clustering-, and LDA-based methods, where the first two methods are unsupervised and the third method is supervised. Experimental results have shown that the proposed method outperforms the unsupervised methods and approaches to the supervised method. Moreover, it is insensitive to different parameter settings of UBM and MBN, which facilitates its practical use.

\section{Acknowledgement}
The author thanks Prof DeLiang Wang for providing the Ohio Computing Center and Dr Ke Hu for helping with the SSC corpus.


\bibliographystyle{IEEEbib}
\bibliography{zxlrefs}

\end{document}